\pdfoutput=1
\documentclass[10pt,twocolumn,letterpaper]{article}

\usepackage{wacv}

\usepackage{times}
\usepackage{epsfig}
\usepackage{graphicx}
\usepackage{amsmath}
\usepackage{amssymb}

\usepackage{array}
\usepackage{mathtools}
\usepackage{xcolor}
\usepackage{algorithm}
\usepackage{algorithmic}
\usepackage{comment}
\usepackage{balance}
\usepackage{float}
\usepackage[accsupp]{axessibility}

\newenvironment{itemizen}
{ \begin{itemize}
    \setlength{\itemsep}{0pt}
    \setlength{\parskip}{0pt}
    \setlength{\parsep}{0pt}     }
{ \end{itemize}                  } 

\newcommand{\xv}{\mathbf{x}}

\newcommand{\bv}{\mathbf{b}}
\newcommand{\R}{\mathbb{R}}
\newcommand{\bmat}[1]{\begin{bmatrix}#1\end{bmatrix}}

\newcommand{\sfactor}{$\sigma$-factor}

\def\wacvPaperID{484} % Enter the WACV Paper ID here

\wacvfinalcopy % *** Uncomment this line for the final submission

\ifwacvfinal
\fi

\ifwacvfinal
\usepackage[breaklinks=true,bookmarks=false]{hyperref}
\else
\usepackage[pagebackref=true,breaklinks=true,colorlinks,bookmarks=false]{hyperref}
\fi

\begin{document}

%%%%%%%%% TITLE
\title{Improving Fractal Pre-training}

\author{Connor Anderson\\
Brigham Young University\\
{\tt\small connor.anderson@byu.edu}
% For a paper whose authors are all at the same institution,
% omit the following lines up until the closing ``}''.
% Additional authors and addresses can be added with ``\and'',
% just like the second author.
% To save space, use either the email address or home page, not both
\and
Ryan Farrell\\
Brigham Young University\\
{\tt\small farrell@cs.byu.edu}
}

\maketitle
\ifwacvfinal
\thispagestyle{empty}
\fi

%%%%%%%%% ABSTRACT
\begin{abstract}
The deep neural networks used in modern computer vision systems require enormous image datasets to train them.  These carefully-curated datasets typically have a million or more images, across a thousand or more distinct categories.  The process of creating and curating such a dataset is a monumental undertaking, demanding extensive effort and labelling expense and necessitating careful navigation of technical and social issues such as label accuracy, copyright ownership, and content bias.

\textbf{What if we had a way to harness the power of large image datasets but with few or none of the major issues and concerns currently faced?}  This paper extends the recent work of Kataoka \etal~\cite{Kataoka2020Pre-trainingImages}, proposing an improved pre-training dataset based on dynamically-generated fractal images.  Challenging issues with large-scale image datasets become points of elegance for fractal pre-training: perfect label accuracy at zero cost; no need to store/transmit large image archives; no privacy/demographic bias/concerns of inappropriate content, as no humans are pictured; limitless supply and diversity of images; and the images are free/open-source.  Perhaps surprisingly, avoiding these difficulties imposes only a small penalty in performance.  Leveraging a newly-proposed pre-training task---multi-instance prediction---our experiments demonstrate that fine-tuning a network pre-trained using fractals attains 92.7-98.1\% of the accuracy of an ImageNet pre-trained network. Our code is publicly available.\footnote{\url{catalys1.github.io/fractal-pretraining/}}

\end{abstract}

\section{Introduction}

One of the leading factors in the improvement of computer vision systems over the last decade has been the access to ever-expanding datasets that can be used for pre-training deep learning models. Nearly all state-of-the-art systems these days have been trained on millions, tens-of-millions, or even hundreds-of-millions of images. Collecting, labeling, managing, and distributing these datasets requires monumental effort---indispensable effort---to achieve the power found in  today's models. However, the list of challenges and concerns around using these datasets is growing as well. Along with technical challenges and high costs, there have been questions of privacy, ownership, inappropriate content, and unfair bias (for example, see~\cite{Birhane2021LargeVision,Steed2020ImageBiases}). Clearly there are complex issues that still need to be overcome, and many of them elude simple solutions.

What if we had a way to harness the power of large image datasets with few or none of the major issues and concerns currently faced? In this paper, we discuss the possibility of using abstract, computer-generated fractal images to pre-train modern computer vision models. We expand on the work of Kataoka et al.~\cite{Kataoka2020Pre-trainingImages}, from whom we take our inspiration. There are several distinct advantages to using fractal images for pre-training:
\begin{itemizen}
    \item Fractals are complex geometric structures that often emerge from a very small set of parameters or equations; thus, as images, they are highly compressible---often a handful of bytes is sufficient to describe them, along with a generic program for rendering them. Therefore, \textbf{the need to store and transmit large datasets of image files can be circumvented}.
    \item Since the data is synthetic, \textbf{we get labels for free}: no massive manual labeling effort is required.
    \item Since fractals are abstract geometric objects, \textbf{there are no issues surrounding the depictions of people}: concerns about privacy, inappropriate content, and biases related to gender, race, or any other human factor can be laid to rest as far as the pre-training data is concerned.
    \item Fractals are ``free and open-source'': they are defined by fairly simple mathematics and anyone can produce the images with only a few dozen lines of code. Thus, there are no issues surrounding copyrights and ownership of images. \textbf{Anyone can use fractal-generated data to train models for any purpose, commercial or otherwise}.
    \item Fractals provide \textbf{a near-limitless supply of diverse images}. Small changes to selected parameters can result in entirely new datasets.
    \item In some cases, fractal images can be very efficient to render. In fact, with the right approach, fractal images can be generated on-the-fly, fast enough to keep up with the throughput of neural network training---even when using fairly large batch sizes and distributed training on multiple GPUs. This both \textbf{eliminates the need to generate and store a fixed set of data} up front, and \textbf{removes the disk-IO bottleneck} that subsequently can become a problem while reading large volumes of data from persistent storage.
\end{itemizen}

A few of the preceding claims are obviously true by virtue of the nature of the data. We show the remaining claims to be true by analysis and experimental evaluation in this paper. The remaining question, then, is how well do fractal-image pre-trained models perform on real-world, natural-image tasks? We show that, while not yet as good as ImageNet pre-training in most cases, the gap is not as large as you might expect.

\begin{figure*}
    \centering
    \includegraphics[width=\linewidth]{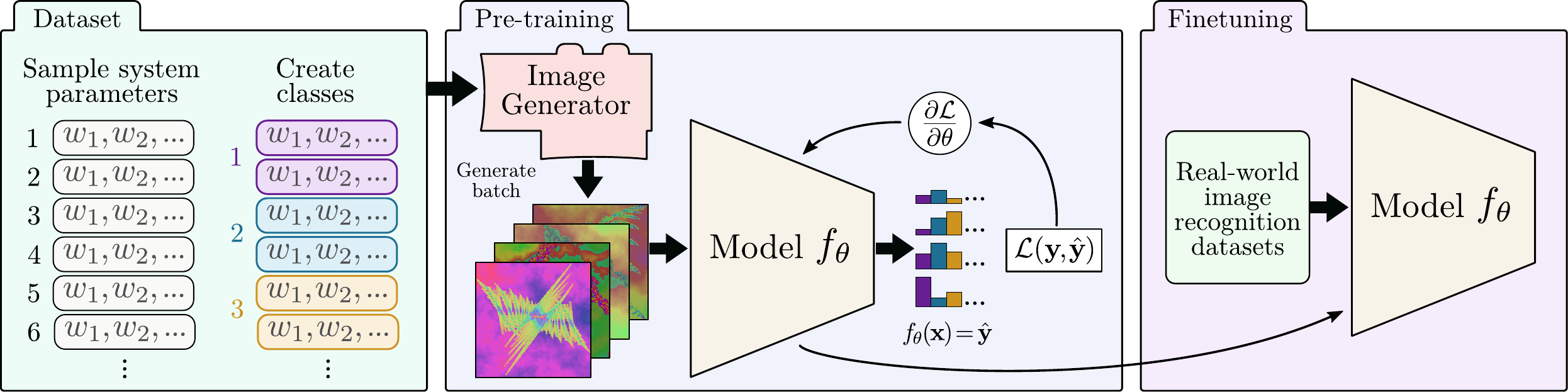}
    \caption{Fractal pre-training. We generate a dataset of IFS codes (fractal parameters), which are used to generate images on-the-fly for pre-training a computer vision model, which can then be fine-tuned for a variety of real-world image recognition tasks.}
    \label{fig:overview}
\end{figure*}

We emphasize that we are not the first to propose pre-training with fractal images. Kataoka et al.~\cite{Kataoka2020Pre-trainingImages} recently introduced the idea, and showed that it was a viable approach. Our work builds on some of the core ideas from their paper---particularly, the use of a large set of randomly sampled affine Iterated Function Systems (IFS) for generating training data. We make several fairly significant deviations from their approach, however, and show that these deviations make a significant difference in the results obtained. Figure~\ref{fig:overview} gives a high-level view of our approach. 

The core contributions of our paper can be summarized as follows:
\begin{itemizen}
    \item We propose a novel, principled approach for sampling IFS codes (see Section~\ref{sec:sampling-ifs}). Our approach leads to highly efficient sampling of large numbers of codes (fractal parameters), simultaneously improving the quality of the resulting fractal images, leading to more effective representation learning.
    \item We introduce large-scale multi-instance (multi-label) prediction as a pre-training task/method (see Section~\ref{sec:tasks}), and show that it is more effective for fractal pre-training than normal multi-class classification, as evaluated on a set of natural-image recognition tasks.
    \item We show that using fractal images generated with color and backgrounds (see Section~\ref{sec:render-fractals}) for pre-training leads to better transfer learning (fine-tuning).
    \item We show that fractal-image pre-training can be quite effective when transferred to tasks with limited training data, such as fine-grained visual categorization and medical image segmentation (see Section~\ref{sec:experiments}).
    \item We show that we can use ``just-in-time'' image generation during training, without ever needing to create and store a database of images beforehand (see Section~\ref{sec:datasets}).
\end{itemizen}

\section{Related Work} \label{sec:related-work}

Over the past decade, ``ImageNet pre-training'' has become an integral part of training computer vision models. The default process is to train a model to perform supervised classification on the 1,000-class ImageNet dataset~\cite{Russakovsky2015ImageNetChallenge} and then fine-tune the model on a different dataset, which has proven to be very effective~\cite{Huh2016WhatLearning, Kornblith2019DoBetter}. Recent work has attempted to probe the limits of this ``supervised classification for pre-training'' approach, showing that huge amounts of data can improve the pre-training transfer performance~\cite{Sun2017RevisitingEra}---even when the labels are only weakly associated with the image content~\cite{Mahajan2018ExploringPretraining}. Large-scale domain-specific datasets, such as iNaturalist~\cite{VanHorn2018TheDataset}, have also proven effective for pre-training~\cite{Cui2018LargeLearning}; large-scale, weakly-labeled data~\cite{Krause2016TheRecognition} has also proven surprisingly effective. Other work has suggested that pre-training isn't the best approach in some domains, such as COCO~\cite{Lin2014MicrosoftContext} object detection~\cite{He2019RethinkingPre-Training,Zoph2020RethinkingSelf-training}; for many problems, particularly those with limited data, however, pre-training provides a substantial performance boost.

%Supervised classification on a large dataset is the method that is most often used for pre-training. However, there have been many alternative unsupervised/self-supervised methods proposed as well---a survey is provided in~\cite{Jing2020Self-supervisedSurvey}. Self-supervised pre-training methods rely on \textit{pretext tasks}, which generally corrupt or augment the images in some way and then attempt to use them to make predictions or recover the originals; examples include colorization~\cite{Zhang2016ColorfulColorization}, jigsaw~\cite{Doersch2015UnsupervisedPrediction,Noroozi2016UnsupervisedPuzzles}, and inpainting~\cite{Pathak2016ContextInpainting}. Recent \textit{contrastive} methods~\cite{Jaiswal2020ALearning}---such as SimCLR~\cite{Chen2020ARepresentations}, MoCo~\cite{He2020MomentumLearning}, and BYOL~\cite{Grill2020BootstrapLearning}---have proved particularly effective. The advantage of self-supervised learning is that it doesn't require image labels; but it does still require a large volume of image data.

For some domains, it can be challenging or impossible to collect and/or annotate enough images to sufficiently pre-train a model. One way to address this issue is to use synthetically generated data~\cite{nikolenko2019synthetic}. Such data can be generated using 3D models~\cite{Tremblay2018TrainingRandomization} or generative models~\cite{bowles2018gan,Triastcyn2019GeneratingLearning}. Usually the data is modeled after natural images and real-world objects.

In contrast to using synthetic data modeled after real-world images, Kataoka~\etal~\cite{Kataoka2020Pre-trainingImages} recently proposed the use of fractal images for pre-training. Fractal images are both synthetic and abstract, though they have some similarity to fractal structures in nature. Fractals have been admired for their visual complexity and beauty, and can be used to create beautiful artwork~\cite{draves2008fractal}; but they have also found practical application in image compression~\cite{fisher2012fractal}, and have even inspired neural network architectures~\cite{Larsson2016FractalNet:Residuals}. Farmer~\cite{farmer2015application} provides a detailed treatment of applications of fractals in computer vision.
Dym~\etal~\cite{DymEtal} contributed a recent study of piecewise-linear functions generated by fractal IFSs and their similarity to those generated by deep neural networks.
In another interesting recent contribution, Marasca~\etal~\cite{MarascaEtal} utilize fractals to assess dataset classification complexity.
Early work used fractal principles for texture segmentation~\cite{kocic1995fractals}, and Kocic~\cite{kocic1995fractals} discussed how fractals could be used to model natural forms.

In this work, we build on~\cite{Kataoka2020Pre-trainingImages} and use fractal images to pre-train visual recognition models.

%\todo{Reviews said that talking about unsupervised methods is irrelevant (probably right), more on fractals.}

\section{Fractal Images} \label{sec:fractal-images}

Fractal images are generally produced by iteration of a simple formula. For example, the well known Mandelbrot and Julia sets can be generated from the equation $z_{k+1}=z_k^2+c$, for $z, c \in \mathbb{C}$, by treating pixel coordinates as points in the complex plane and iterating until $z$ ``escapes'' toward infinity or remains bounded for some number of iterations. Information about whether or not the point escaped, and how long it took to do so, can be used to color the pixels, revealing rich complexity.

An Iterated Function System (IFS) can also be used to generate fractal images. An IFS consists of a set of two or more functions and an associated set of probabilities. The set of functions, which we refer to as a \textit{system} or a \textit{code}, have an associated \textit{attractor}---a set of points with a particular geometric structure---such that iterative application of the functions in the system will bring points in the associated space onto the attractor. Sec.~\ref{sec:render-fractals} describes how fractal images can be rendered from IFS codes. These images exhibit complex patterns and self-similarity.% \todo{See figure with examples?}

As proposed in~\cite{Kataoka2020Pre-trainingImages}, we use affine Iterated Function Systems to create a dataset of fractal images for pre-training computer vision models. In an affine IFS, the functions in the system are affine transformations: they consist of a linear function, represented by a matrix $A$, and a translation vector $\mathbf{b}$, so that $w(x) = A\mathbf{x} + \mathbf{b}$. Affine functions have several advantages: particularly, it is easy to evaluate whether they are contractive functions (which is generally necessary for IFS) and they are simple and fast to evaluate numerically.

\paragraph{Iterated Function Systems} \label{iterated-function-systems}

\newcommand{\attractor}{\mathcal{A}_{\mathcal{S}}}
\newcommand{\iterpoints}{\hat{\mathcal{A}}}

We now provide a more formal definition of Iterated Function Systems. An IFS code $\mathcal{S}$ defined on a complete metric space $\mathcal{X}$ (we will assume that the metric space is $\mathcal{X}=(\R^2, \|\cdot\|_2)$) is a set of transformations $w_i : \mathcal{X} \rightarrow \mathcal{X}$ and their associated probabilities $p_i$:
\begin{equation}
    \mathcal{S}=\{(w_i, p_i) : i=1,2,\dots,N\}, \label{eq:ifs-codes}
\end{equation}
which satisfy the average contractivity condition
\begin{equation}
    \prod_{i=1}^N s_i^{p_i} < 1, \label{eq:average-contractivity}
\end{equation}
where $s_i$ is the Lipschitz constant for $w_i$. The \textit{attractor} $\attractor$ is a unique geometric structure~\cite{Barnsley1988AEncoding}, a subset of $\mathcal{X}$ defined by $\mathcal{S}$. The shape of $\attractor$ depends on the functions $w_i$ and not the probabilities $p_i$; however, the $p_i$ do affect the distribution of points across $\attractor$\footnote{See~\cite{Barnsley1988FractalImages} for a more thorough introduction to Iterated Function Systems.}. We choose $p_i \propto \left|\det{A_i}\right|$, as done in \cite{Kataoka2020Pre-trainingImages}. We provide a visual comparison between determinant-proportional and uniform $p_i$ in Appendix~\ref{sec:system-probabilities}.

\subsection{Rendering Fractal Images} \label{sec:render-fractals}

We can render an approximation of $\attractor$ to obtain a fractal image. The random iteration method, or ``chaos game'', can be used to to generate a subset of $\attractor$ as follows: choose an initial point $\xv_0 \in \mathcal{X}$; randomly choose $w_i$ with probability $p_i$ and apply it to $\xv_0$ to get $\xv_1=w_i(\xv_0)$; repeat this process for a sufficiently large number of iterations $K$ to get the set of points $\iterpoints = \{\xv_0, \xv_1,\dots, \xv_K\} \subseteq \attractor$. The larger $K$ is, the closer the approximation will be to $\attractor$. 

The next step is to render the points in $\iterpoints$ to an image $X$. We map a rectangular region $\mathcal{R} \in \R^2$, nominally defined by the min and max $x$ and $y$ values in $\iterpoints$, to an $M \times M$ pixel grid. Pixels can be rendered as binary values, indicating that at least one point in $\iterpoints$ maps to that pixel; or they can be rendered as continuous values, indicating the density of points that map to each pixel.

\begin{figure}
    \centering
    \includegraphics[width=\linewidth]{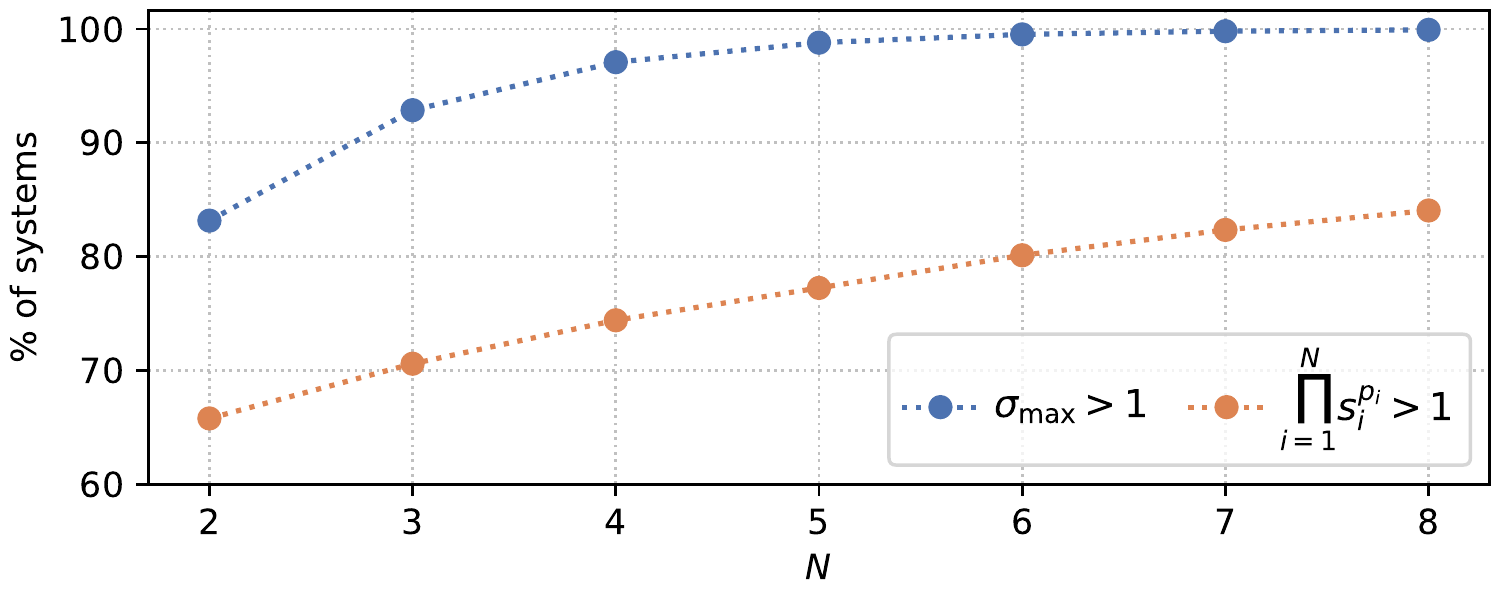}
    \caption{\small For systems with $N=2,\dots,8$ transforms, we show the percentage of systems (out of 100,000 randomly sampled) that have their largest singular value greater than $1$ (in red), and also those which violate average contractivity (in blue: $p_i\propto \det\left(A_i\right)$, where $A_i$ is the linear part of the affine transform).}
    \label{fig:unconstrained-systems}
\end{figure}

\begin{figure*}
    \centering
    \includegraphics[width=\linewidth]{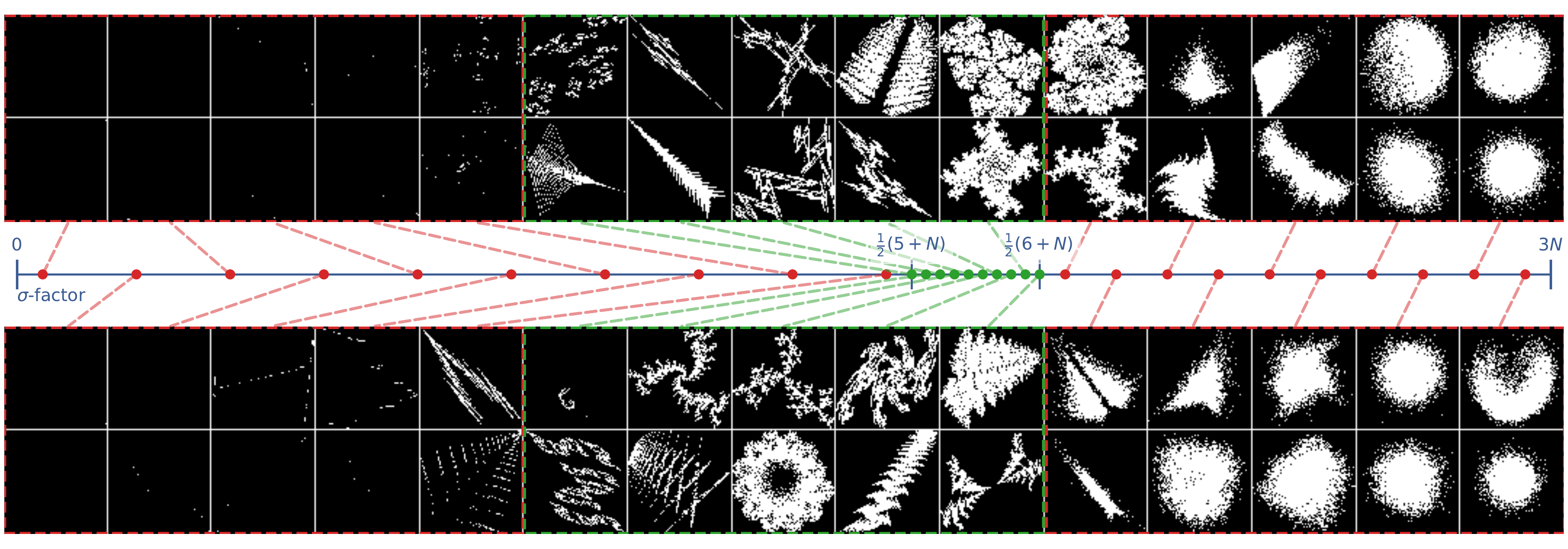}
    \caption{\small Fractal systems by \sfactor~(see Eq.~\ref{eq:sigma-factor}). IFS codes with a \sfactor\ in the range $[\frac{1}{2}(5+N), \frac{1}{2}(6+N)]$ (where $N$ is the size of the system) tend to yield images with good geometry, while many codes with a \sfactor\ outside this range yield images with degenerate geometry. Images were generated randomly at each \sfactor, with $N=2$}
    \label{fig:sigma-factor-images}
\end{figure*}

\subsection{Sampling Iterated Function Systems} \label{sec:sampling-ifs}

%\todo{I don't know if I like the term "good geometry", but I'm not sure what else to call it right now.}

% \noindent \textbf{Note on notation:} Throughout the paper, we use $U(a, b)$ to mean a continuous uniform distribution on the interval $[a, b]$, and $U(\{\cdot\})$ to mean a discrete uniform distribution over elements of the set $\{\cdot\}$.

\subsubsection{What Makes a ``Good'' IFS Code?} \label{sec:good-ifs}

So far we have defined Iterated Function Systems and how we use them to render fractal images. We now turn our attention to the question of how to get the IFS codes in the first place. Let $N = |\mathcal{S}|$, the number of functions in the code $\mathcal{S}$. In~\cite{Kataoka2020Pre-trainingImages}, they choose codes by sampling $N\thicksim U(\{2,3,\dots,8\})$\footnote{Throughout the paper, we use $U(a, b)$ to mean a continuous uniform distribution on the interval $[a, b]$, and $U(\{\cdot\})$ to mean a discrete uniform distribution over elements of the set $\{\cdot\}$.}, and then sampling the six values $(A_k, \bv_k)$ for each of the $N$ affine transformations from $U(-1,1)$. Repeating this process thousands of times produces a dataset of IFS codes. Each code can then be treated as a ``class'', for the purpose of doing supervised multi-class pre-training.

There are a few problems with the sampling approach taken in~\cite{Kataoka2020Pre-trainingImages}. First, in order to be a true IFS, the system must be a contraction (Eq.~\ref{eq:average-contractivity}). Sampling random transforms with values between $-1$ and $1$ does not guarantee a contraction.  In fact, the majority of codes thus sampled will not be. To demonstrate this, we sampled 100,000 random systems with parameters in $U(-1,1)$ for each of $N=2,\dots,8$. Figure~\ref{fig:unconstrained-systems} shows (i) in blue, the percentage of those systems that had \emph{at least} one singular value greater than $1$---an affine transform must have singular values less than $1$ to be a contraction, as we describe shortly---and (ii) in red, the percentage that violate the average contractivity condition. Clearly, a naive sampling approach is quite inefficient, as the majority of systems will not be contractions, and when a system is not contractive it will, under iteration, produce sequences that diverge to infinity. Such sequences cause numerical difficulties and yield unsatisfactory images. 

The second problem is that even when a system is a contraction, it might not produce fractals with ``good'' geometric properties. What do we mean by ``good'' geometric properties? Consider the fractal images shown in Figure~\ref{fig:sigma-factor-images}. Those on the left are very sparse, consisting of mostly blank space and perhaps a few small structures (note that this is similar to the idea of ``filling rate'' as discussed in~\cite{Kataoka2020Pre-trainingImages}). Those on the right look like blurry smudges. The ones in the middle, however, contain complex and varied structure. This set is the most visually interesting; we hypothesize (and experimentally validate in Section \ref{sec:exper-ablation}) that it is the most useful for representation learning.

\subsubsection{Effectively Sampling IFS Codes} \label{sec:sampling-algos}

We will now describe an approach to sampling IFS codes that addresses the two concerns just raised (contractivity and good geometry). Our approach is based on the Singular Value Decomposition (SVD) of $A_k$, the linear part of the affine transform. First, in order to ensure that an IFS is contractive, it is sufficient to ensure that each function $w_i$ is a contraction; that is, it satisfies
\begin{equation}
    \|w_i(\xv_1) - w_i(\xv_2)\| \leq \|\xv_1 - \xv_2\|,\; \forall\ \xv_1, \xv_2 \in \mathcal{X}
\end{equation}
For affine functions $w_i(\xv) : \mathbb{R}^2\!\rightarrow \mathbb{R}^2 = A\mathbf{x} + \mathbf{b}$, we require
\begin{equation}
\begin{aligned}
    &{} & \|A\xv_1+\bv - A\xv_2-\bv\|_2 &\leq \|\xv_1-\xv_2\|_2 \\
    &\Rightarrow & \frac{\|A(\xv_1-\xv_2)\|_2}{\|\xv_1-\xv_2\|_2} &\leq 1 \label{eq:contract} \\
    &\Rightarrow & \sigma_{\max}(A) &\leq 1
\end{aligned}
\end{equation}
where $\sigma_{\max}(A)$ denotes the maximum singular value of $A$~\footnote{The final line of Eq.~\ref{eq:contract} follows from the definition of the spectral norm, or $l_2$ operator norm: $\|A\|_2 = \sup_{\xv\neq0}\frac{\|A\xv\|}{\|\xv\|}=\sigma_{\max}(A)$}. Thus, it is sufficient to ensure that the singular values of $A$ are less than $1$, which we can do by construction. Recall that by the Singular Value Decomposition, $A=U\Sigma V^T$, where $U$ and $V$ are orthogonal matrices and $\Sigma$ is a diagonal matrix containing the singular values of $A$, $\sigma_1$ and $\sigma_2$, ordered by decreasing magnitude. Since $U$ and $V$ are orthogonal, they act as rotation matrices (with possible reflection, \ie the determinant can be $\pm 1$). 
Let $U=\mathcal{R}_{\theta}$ be a rotation by angle $\theta$, and let $V^T=\mathcal{R}_{\phi}$ be a rotation by angle $\phi$. Let $D$ be a diagonal matrix with diagonal elements $d_1,d_2 \in \{-1,1\}$. Then $A = U\Sigma V^T = \mathcal{R}_{\theta}\Sigma\mathcal{R}_{\phi}D$. We can sample $A$ by appropriately sampling $(\theta, \phi, \sigma_1, \sigma_2, d_1, d_2)$, composing the corresponding matrices, as above, and then multiplying them together to obtain $A$. By sampling $\sigma_1$ and $\sigma_2$ in the range $(0,1)$, we ensure that the system is a contraction.

Sampling the SVD parameters directly guarantees a contraction mapping, however, it does not address the question of good geometry. It's not immediately clear what the relationship between the fractal geometry and the system parameters is, nor whether there is a simple and concise relationship at all.
% \todo{[Seems a little informal:] It seemed likely to us that the singular values of the system were an important factor though}.
Intuition, however, hinted that the singular values might play an important role.
The magnitudes of the singular values define how quickly an affine contraction map converges to its fixed point under iteration; small singular values would lead to quick collapse, while singular values near $1$ would be more likely to result in ``wandering'' trajectories which don't converge to a clear geometric structure.
Closer investigation revealed that there is indeed a correlation between some property of the singular values and whether the resulting fractal possesses good geometry. To probe this relationship, we labeled by hand several hundred size-2 systems according to whether they had good geometry or not (subjectively), and fit a linear Support Vector Machine (SVM) classifier to the labels using the singular values of the system as features. The SVM was able to distinguish between the systems with nearly perfect accuracy. We repeated this experiment several times for different system sizes. As we then investigated the decision boundaries learned by the classifiers, we discovered a simple and general pattern.
Let $\sigma_{i,1}$ and $\sigma_{i,2}$ be the singular values for $A_i$, the $i$th function in the system, and let $\xv_\sigma=\bmat{\sigma_{1,1} & \sigma_{1,2} & \dots & \sigma_{N,1} & \sigma_{N,2}}^T$, and $\mathbf{w}_\sigma=\bmat{1 & 2 & \dots & 1 & 2}^T$. We find that a large majority of the systems with good geometry satisfy $\alpha_{l} \leq \mathbf{w}^T_\sigma \xv_\sigma \leq \alpha_u$; in other words, confining the weighted sum of a system's singular values
\begin{equation} \label{eq:sigma-factor}
    \alpha = \sum_{i=1}^N(\sigma_{i,1}+2\sigma_{i,2}) 
\end{equation}
to the appropriate range $(\alpha_l, \alpha_u)$ will produce systems with mostly good geometry, while systems outside of that range tend to have less desirable geometry. We refer to the quantity $\alpha$ in Eq.~\ref{eq:sigma-factor} as the \sfactor\ of the system. The appropriate range $(\alpha_l, \alpha_u)$ depends on $N$, the size of the system; but empirically, we discovered that setting $\alpha_l = \tfrac{1}{2}(5+N)$ and $\alpha_u = \tfrac{1}{2}(6+N)$ works very well for $N=2,\dots,8$---although a wider range might also work. Figure~\ref{fig:sigma-factor-images} shows the effect of sampling images at different \sfactor s.

We now know that we can confidently tell whether a system will have good geometry by looking at its \sfactor. We next describe an algorithm to randomly sample a set of singular values, $(\sigma_{1,1},\sigma_{1,2},\dots,\sigma_{N,1},\sigma_{N,2})$ that satisfy the conditions
\begin{equation} \label{eq:sv-bounds}
    0 \leq \sigma_{i,2} \leq \sigma_{i,1} \leq 1
\end{equation}
and Eq.~\ref{eq:sigma-factor} for some $0 \leq \alpha \leq 3N$. We take an iterative approach, sampling one singular value at a time and updating the constraints on the next one accordingly. We start with $\sigma_{1,1}$; it could achieve its smallest possible value if every other were maximized. Assume that every other singular value was maximized according to Eq.~\ref{eq:sv-bounds}, then we have $\alpha = \sigma_{1,1} + 2\sigma_{1,2} + 3(N-1)$, and the lower bound on $\sigma_{1,1}$ is $\max(0, \tfrac{1}{3}(\alpha-3(N-1)))$. Similarly, $\sigma_{1,1}$ could achieve its maximum value when all others are minimized, so we set them to $0$ and get that the upper bound on $\sigma_{1,1}$ is $\min(1, \alpha)$. We then sample $\sigma_{1,1}$ uniformly according to the bounds just established, and it becomes a constant in all further bounds calculations. We follow this same process for all but the last two singular values, calculating upper and lower bounds for---and then sampling---each singular value in turn, and updating the constraints on future values. For the last pair, it is more convenient to first sample $\sigma_{N,2}$, at which point $\sigma_{N,1}$ becomes fixed in order to satisfy Eq.~\ref{eq:sigma-factor}. %The full algorithm is given in Alg.~\ref{alg:sample-svs}.

The sampling constraints given by Eqs.~\ref{eq:sigma-factor} and~\ref{eq:sv-bounds} lead to a problem that resembles a Linear Program, except that instead of trying to find a minimizing point, we need to sample a point on the surface of the resulting $2N$-polytope. The algorithm described above does this by iteratively sampling a value independently in each dimension, restricting the available sampling volume for the remaining dimensions. This approach does not necessarily sample points uniformly across the volume, but it is fast and should be sufficient to sample a diverse set of IFS codes.

We now have a process, described formally as \texttt{sample-svs} in Alg.~\ref{alg:sample-svs} (in Supplementary Material, Appendix~\ref{sec:formal-algorithms}), for sampling singular values so that the resulting systems exhibit good geometry. Our algorithm using this process to sample IFS codes via SVD is described as \texttt{sample-system} in Alg.~\ref{alg:sample-system} (in Supplementary Material, Appendix~\ref{sec:formal-algorithms}).

% \todo{Comment on similarity to SIMPLEX on Linear Program, and how the process as outlined won't sample the ``good'' space uniformly, but hopefully does an adequate job.  Actually sampling uniformly is future work.}

\section{Pre-training Procedures}

\subsection{Pre-training Tasks} \label{sec:tasks}

We are focused on using fractal images to learn good representations which will benefit natural-image recognition tasks via fine-tuning. There are several pre-training tasks that could be used to learn these representations. We could use unsupervised/self-supervised pre-training strategies as described earlier in related work (Section \ref{sec:related-work}), however, these seem less compelling when labels are both accurate and abundant.
Supervised multi-class classification is routinely used for pre-training on ImageNet, and is the pre-training task adopted in~\cite{Kataoka2020Pre-trainingImages}.
% \todo{REWRITE AFTER REMOVING UNSUPERVISED} %There are also a variety of unsupervised feature-learning approaches that have been used in the natural-image domain (see Section~\ref{sec:related-work}), which might also prove effective with fractals. We explore both of these approaches in our experiments.
We too utilize this widely-accepted approach; we additionally propose a new pre-training method which is uniquely suited to synthetically-generated data such as fractal images, a task which we call \textit{multi-instance prediction}. Multi-instance prediction is a type of large-scale multi-label classification, where each image may contain examples of multiple classes.
We describe each of these approaches in greater detail below.

\subsubsection{Multi-class Classification} \label{sec:multi-class-classification}

For multi-class classification, we follow the same basic approach taken in~\cite{Kataoka2020Pre-trainingImages}. We choose a fixed number of classes $C$, and assign IFS codes to each class. We use the standard cross-entropy objective function to train the model to predict the corresponding class for each image.

\noindent \textbf{Assigning classes to IFS codes}\quad Kataoka \etal~\cite{Kataoka2020Pre-trainingImages} assign each IFS code its own class label, and then augment each class by scaling each of the six parameters in $(A_k, \bv_k)$ individually, essentially yielding a set of codes for each class. In principle, these codes will be related, since there is a smoothness to the space of fractals defined by IFS. Small perturbations to the parameters can still yield large differences in the resulting fractals, however (see example in Appendix~\ref{sec:data-examples}); and simply scaling the parameters of the affine transformations may additionally cause the singular values of the system to become too large or too small, producing sparse or degenerate images.
% \todo{In fact, we can see this in the official FractalDB dataset released by the authors.}

Our approach is to sample more systems according to Algorithm~\ref{alg:sample-system}, and assign a single class label to a \emph{group} of IFS codes. In our experiments, we show that this approach outperforms the parameter-scaling method of~\cite{Kataoka2020Pre-trainingImages}. However, we also experiment with some parameter augmentation methods and find that they can still help performance.

%
%\subsubsection{Self-supervised Learning}
%\todo{TODO}

\subsubsection{Multi-instance Prediction} \label{sec:multi-instance-prediction}

Multi-instance classification is a supervised classification task; like multi-class classification, we define a fixed number of classes $C$ and assign one or more IFS codes to each class. But unlike multi-class classification, the images we use in the multi-instance setting may contain multiple fractal instances from multiple classes---hence ``multi-instance''. During training, the model performs multi-label prediction, trying to determine the presence or absence of each of the $C$ classes in each image. In other words, each class can be considered as an attribute, and the model tries to predict which attributes are present.

Multi-instance prediction is significantly more challenging than multi-class classification, as evidenced in our experiments. Each image contains a variable number of fractals, yielding a vast space of possible image configurations. In training, the model must learn to pay attention to each fractal ``attribute'' that is present. Our experiments show that this added task complexity leads to pre-trained features that generalize significantly better for downstream tasks.

The images for training multi-instance classification models looks different than for regular multi-class, the process for generating them is different, and there are some special considerations that need to be accounted for. We discuss this in Section~\ref{sec:multi-instance-images}.

We train multi-instance models using a binary cross-entropy loss, averaged across all the classes. Since the number of positive examples in each image is so small compared to number of classes (e.g. 5 out of 1000), we apply a large weight to the positive examples to balance the loss. For instance, when using 1000 classes and a maximum of 5 instances per image, we multiply the loss of the positive classes in each image by 200. Without applying this weighting factor, the model fails to learn.

\subsection{Pre-training Datasets} \label{sec:datasets}

Each pre-training task operates on different types of images: single-fractal images for multi-class classification, and multi-fractal images for multi-instance prediction. For both tasks, images are not generated or stored up-front; images are generated ``just-in-time'', as needed during training.

With the correct procedure, we are able to \emph{generate} all images ``on the fly'' as they're needed for training. This is significant, as we circumvent the typical need to store or transmit a huge quantity of data. The entire dataset can be generated from the set of IFS codes, which can be stored in tens or hundreds of megabytes (depending on the number and size of the systems). For example, an ImageNet-sized (1.28M images) fractal dataset requires only 184.5MB to store its parameters instead of the 150GB of storage needed to store ImageNet (ILSVRC 20212). 
We leverage an efficient Numba~\cite{Lam2015Numba:Compiler} implementation, along with a \textit{rendering cache} of recently generated images, in order to achieve the necessary throughput during training.
Please see Appendix \ref{sec:jit-image-generation} in the Supplementary Material for details.

%%% RENDERED FRACTAL IMAGES FIGURE %%%
\begin{figure}
    \centering
    \includegraphics[width=\linewidth]{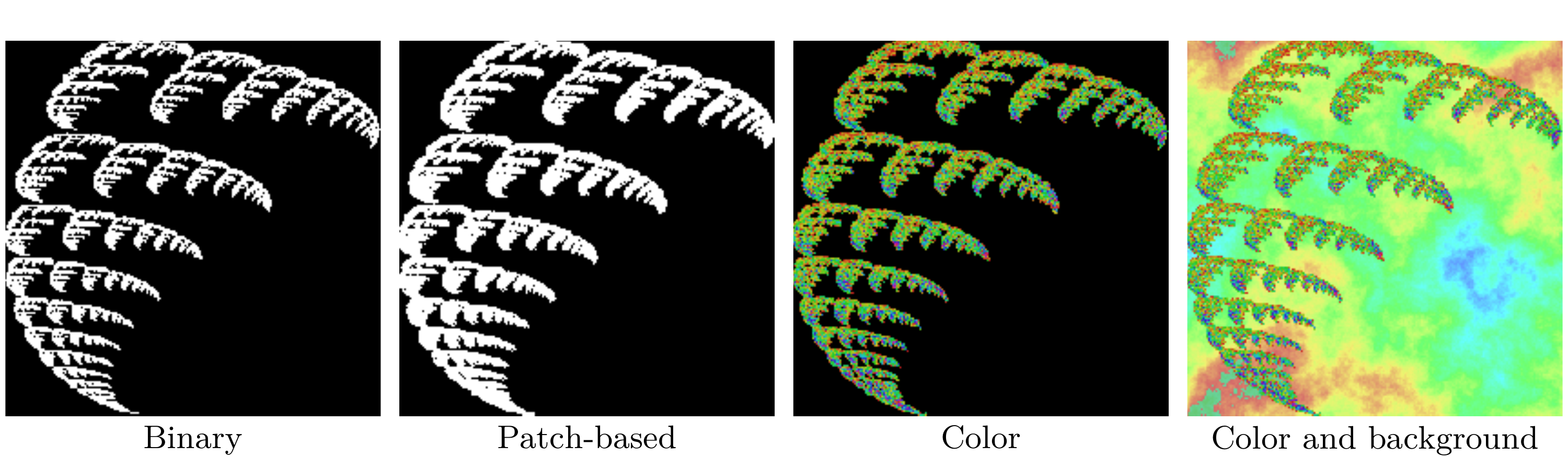}
    \caption{\small Rendered fractal images.}
    \label{fig:rendered-fractals}
\end{figure}
%%%%%%%%%%%%%%%%%%%%%%%%%%%%%%%%%%%%%%

\begin{figure}
    \centering
    \includegraphics[width=\linewidth]{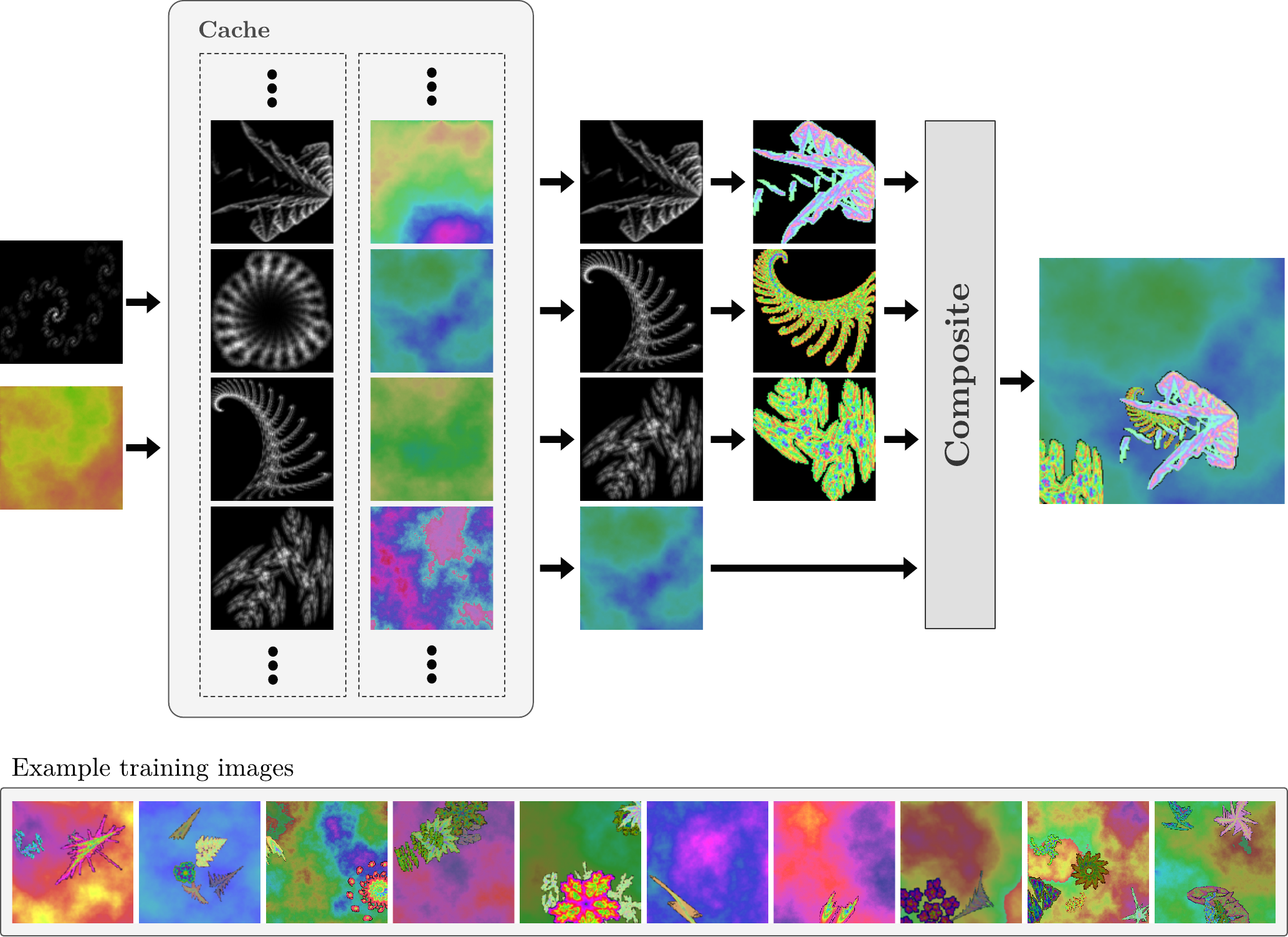}
    \caption{\small Rendering multi-instance images. A cache of fractals and backgrounds is regularly updated with new samples. Each training image is composed of a random selection of fractals from the cache, randomly augmented and composited on top of a random background image. Best viewed digitally, zoom in for details.}
    \label{fig:multi-instance-images}
\end{figure}

\subsubsection{Single-instance Images} \label{sec:single-instance-images}

To generate a given fractal image, we start with the process described in Section~\ref{sec:render-fractals}. We then add three additional elements: patch-based rendering, as described in~\cite{Kataoka2020Pre-trainingImages}; adding color to the fractal; and adding a randomly generated background using the ``diamond-square'' algorithm~\cite{Fournier1982ComputerModels}. Full details are provided in Appendix \ref{sec:rendering-details} in the Supplementary Material. In addition, we randomly scale and translate the region $\mathcal{R}$ (see Section~\ref{sec:render-fractals}), which scales and translates the resulting fractal. We also apply random flips and 90$^{\circ}$ rotations, and random Gaussian blur to the final image.

\begin{figure*}[t]
    \centering
    \includegraphics[width=\linewidth]{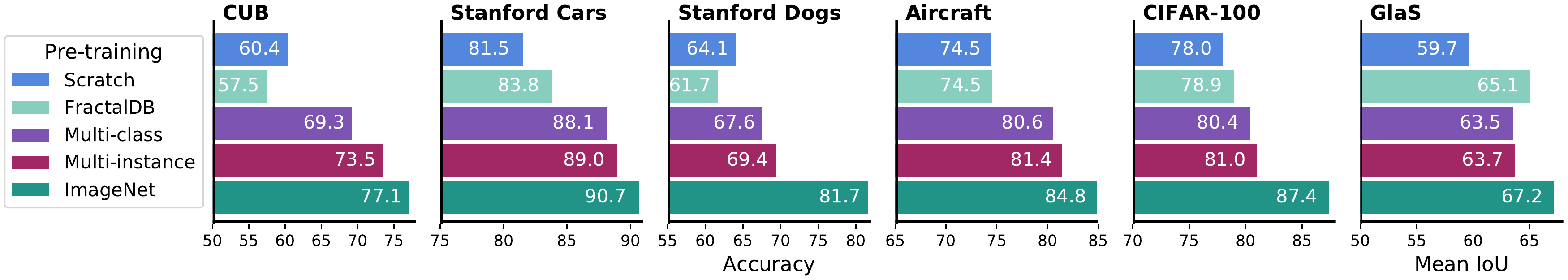}
    \caption{\small Fine-tuning evaluation results (classification accuracy) using different pre-training methods.  The five datasets to the left are image classification datasets; the rightmost dataset, GlaS, is a medical image segmentation dataset.}
    \label{fig:finetune-results}
\end{figure*}

\subsubsection{Multi-instance Images} \label{sec:multi-instance-images}

We create multi-instance images by compositing one or more fractals and a background into a single image. For each image, we randomly sample the number of fractals, $n$, uniformly from $\{1,\dots,N_{\max}\}$. Then we randomly sample $n$ classes and generate their fractal images; we generate the fractals at a lower resolution (such as 128 instead of 224), for efficiency and because of how we composite them. We do not apply scaling or translation at this stage. We also generate a random background. In our experiments, we set $n_{\max}=5$, to produce images which aren't overly cluttered. 

Once we have the $n$ fractals and a background, we composite them together. We randomly rescale each fractal and add it to the image at a random location. Fractals may end up partially occluded by other fractals, or partially outside the image, resulting in complex, varied, and challenging images for recognition (see Figure~\ref{fig:multi-instance-images}, bottom).

Rendering multiple fractals for every training image will almost certainly be too slow; in this case, a rendering cache becomes particularly useful. Every $k_p$ training images (we set $k_p=2$), we generate a new grayscale fractal image and new background, and update the cache. To generate a training image, we choose $n$ random fractals and a background from the cache; we randomly flip and colorize each fractal, and add it to the background image as described previously. This allows us to create multi-instance images with roughly the same computation as in the single-instance case. This process is illustrated in Figure~\ref{fig:multi-instance-images}.

\section{Experiments} \label{sec:experiments}

% Our basic fractal pre-training dataset consists of 50,000 IFS codes grouped into 1,000 classes. 
Our basic fractal pre-training dataset consists of 1,000 IFS codes, each representing a class.
The IFS codes are sampled uniformly for $N \in \{2,3,4\}$, and the parameters are sampled as described in Section~\ref{sec:sampling-algos}. We also employ a parameter augmentation method, which randomly selects one of the transforms $(A_k,\bv_k)$ in the system and scales it by a factor $\gamma \thicksim U(0.8, 1.1)$ (while making sure not to overflow the singular values) to get $(\gamma A_k, \gamma \bv_k)$. We found this to be more effective than the other augmentation methods we explored and plan to investigate why in future work.

\noindent\textbf{Note}: In the conference version of this paper, we stated that we used 50,000 IFS codes grouped into 1,000 classes. A bug in the code made this untrue, and only 1,000 IFS codes were ever actually used, despite our configuration files. We became aware of this after the camera-ready submission. In this updated version, we have removed the erroneous results that were meant to probe the effect of the number of systems versus parameter augmentation, as they didn't actually test what we intended. The other results are still valid, as they weren't based on assumptions about the number of systems per class.

\subsubsection*{Setup/Implementation}

Our experiments use a ResNet50~\cite{He2016DeepRecognition} CNN model. We pre-train each model for 90 epochs, with 1,000,000 training samples per epoch, and with an image resolution of $224\times224$. Most models are pre-trained using 8 NVIDIA GTX 1080 Ti GPUs, with a total batch size of 512. Some of the ablations were run using 4 Tesla P100 GPUs and a batch size of 256.

We evaluate the effectiveness of the pre-trained representations by fine-tuning on several different tasks. For image classification, we use CUB-2011~\cite{WahCUB_200_2011}, Stanford Cars~\cite{KrauseStarkDengFei-Fei_3DRR2013}, Stanford Dogs~\cite{KhoslaJYF_CVPRW2011}, FGVC Aircraft~\cite{MajiRKBV_arXiv2013} and CIFAR-100~\cite{krizhevsky2009learning}. We also fine-tune models for medical image segmentation on the GlaS dataset~\cite{Sirinukunwattana2017GlandContest}.  For classification, we fine-tune for 150 epochs with a batch size of 96---we found that longer fine-tuning led to better performance. For segmentation, we fine-tune for 90 epochs with a batch size of 8.

We compare our proposed methods with three baselines: training from scratch (no pre-training); fine-tuning from ImageNet~\cite{Russakovsky2015ImageNetChallenge} pre-trained weights (available through PyTorch~\cite{Paszke2019PyTorch:Library}); and, fine-tuning from FractalDB~\cite{Kataoka2020Pre-trainingImages} pre-trained weights, which we obtained using the dataset and code made available by the authors\footnote{hirokatsukataoka16.github.io/Pretraining-without-Natural-Images}.

Our experiments use PyTorch~\cite{Paszke2019PyTorch:Library} and the PyTorch-Lightning framework~\cite{falcon2019pytorch}, with Hydra~\cite{Yadan2019Hydra} for configuration. To facilitate reproducible research, all code and configuration files will be made publicly available.

\subsection{Fine-tuning Results}

Fig.~\ref{fig:finetune-results} shows the fine-tuning performance on each evaluation task using different pre-training methods, along with training from scratch. Fractal pre-training provides a clear and consistent boost over both training from scratch as well as pre-training with FractalDB. Using multi-instance prediction as the pre-training task is also consistently better than using multi-class classification. In fact, multi-instance prediction models can provide more than 90\% of the accuracy achieved by ImageNet pre-training---and in some cases, such as for Stanford Cars, the model obtains over 98\% of the ImageNet performance.

\subsection{Ablation Experiments} \label{sec:exper-ablation}

\begin{figure}
    \centering
    \includegraphics[width=\linewidth]{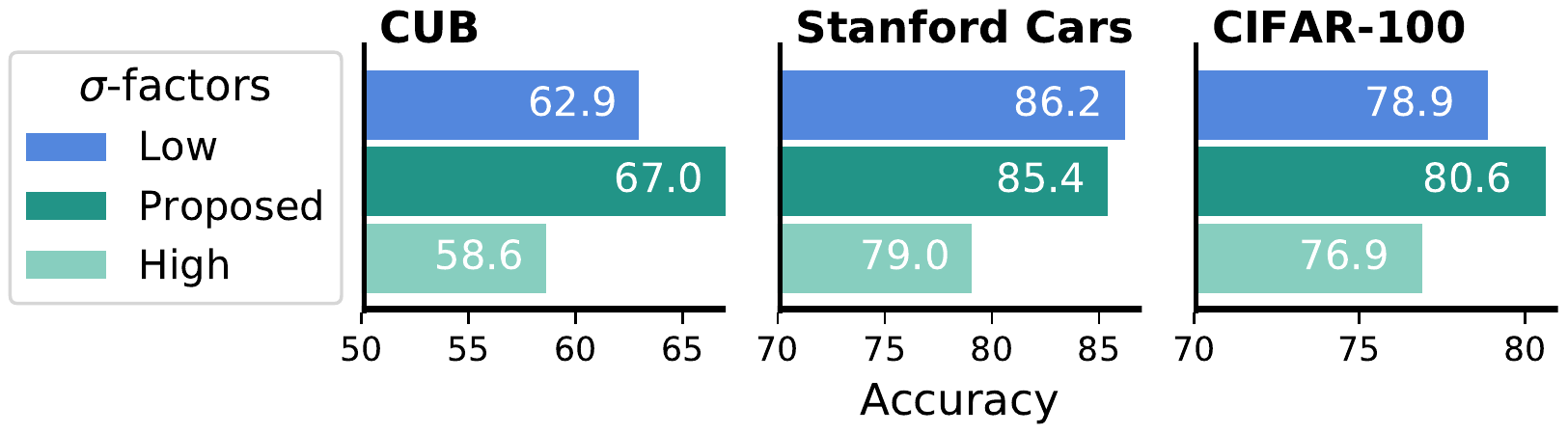}
    \caption{\small Performance from pre-training on systems with different \sfactor\ ranges: (\textbf{low}) $[1, \frac{1}{2}(5+N)]$;  (\textbf{proposed}) $[\frac{1}{2}(5+N), \frac{1}{2}(6+N)]$; (\textbf{high}) $[3N\!-\!1, 3N]$.}
    \label{fig:sfactor-results}
\end{figure}

This section provides ablation experiment results isolating the impact of different parts of the proposed pre-training method. Additional results can be found in the Appendix.

\noindent\textbf{Impact of \sfactor s}\quad First, we consider the effects of using different ranges of \sfactor s.
Figure~\ref{fig:sfactor-results} shows the results of pre-trainng models using three different \sfactor\ ranges (see figure caption for details) and evaluating performance when fine-tuning for the CUB, Stanford Cars and CIFAR-100 datasets.  
Across all three datasets, the \texttt{high} \sfactor s perform poorly relative to the \texttt{proposed} ``good'' geometry range.  Similarly, the proposed \sfactor\ range outperforms the \texttt{low} \sfactor s model on two of the three datasets: CUB and CIFAR-100.

\section{Conclusion}

We have shown that by carefully designing the processes for sampling and rendering affine IFSs, fractal pre-training can yield strong representations that are useful for real-world image recognition tasks. We have also proposed a pre-training task---multi-instance prediction---which greatly improves over multi-class classification as a pre-training task. Finally, we have shown that the fractal images used for pre-training can be generated ``on-the-fly'' in real-time during training, removing the need to generate, store, or transmit a large volume of data.

% In the conclusion, you should restate the thesis and show how it has been developed through the body of the paper. Briefly summarize the key arguments made in the body, showing how each of them contributes to proving your thesis

\noindent
\textbf{Acknowledgments}
This work was supported by the National Science Foundation under Grant No. IIS1651832. We gratefully acknowledge the support of NVIDIA for their donation
of multiple GPUs used in this research.

%\balance

{\small
\bibliographystyle{ieee_fullname}
\bibliography{egbib, ConnorMendeley, added_refs}
}

\appendix

\section*{\centering \LARGE Appendix}
\vspace{0.15in}

\section{Algorithms} \label{sec:formal-algorithms}

Described in the main paper in Section \ref{sec:sampling-algos}, we here provide precise descriptions for the \texttt{sample-svs} and \texttt{sample-system} algorithms, respectively in Algs. \ref{alg:sample-svs} and \ref{alg:sample-system}. We use these algorithms in our experiments to sample the IFS codes used in our fractal dataset.

\begin{algorithm}[t]
\small
\newcommand{\s}{\sigma}
\caption{\texttt{sample-svs($N$,$\alpha$)}: Sample singular values.}
\begin{algorithmic}[1] \label{alg:sample-svs}
    \renewcommand{\algorithmicrequire}{\textbf{Input:}}
    \renewcommand{\algorithmicensure}{\textbf{Output:}}
    \REQUIRE $N \ge 2$, the size of the system; and $0 \le \alpha \le 3N$, the target \sfactor\ of the system.
    \ENSURE  $\Sigma$, the $N\times2$ array of singular values, satisfying 
        $0 \le \Sigma_{k,2} \le \Sigma_{k,1} \le 1\ (\forall\ k=1,\dots,N)$ and
        $\sum_{i=1}^{N}(\Sigma_{i,1}+2\Sigma_{i,2}) = \alpha$.
    \STATE \textbf{Initialize:} $\Sigma\gets \mathbf{0}^{N \times 2}$, the array of singular values
    \STATE \textbf{Initialize:} $b_l\gets \alpha-3N+3$, sampling lower bound 
    \STATE \textbf{Initialize:} $b_u\gets \alpha$, sampling upper bound
    \FOR{$k=1$ to $N-1$}
        \STATE Sample $\s_{k,1} \thicksim U(\max(0,\frac{1}{3}b_l), \min(1, b_u))$
        \STATE Update $b_l\gets b_l-\s_{k,1}$ and $b_u\gets b_u-\s_{k,1}$
        \STATE Sample $\s_{k,2} \thicksim U(\max(0,\frac{1}{2}b_l), \min(\s_{k,1}, \frac{1}{2}b_u))$
        \STATE Update $b_l\gets b_l-2\s_{k,2}+3$ and $b_u\gets b_u-2\s_{k,2}$
        \STATE Update $\Sigma_{k,1}\gets \s_{k,1}$ and $\Sigma_{k,2}\gets \s_{k,2}$
    \ENDFOR
    \STATE \COMMENT{Note the use of $b_u$ in both places below}
    \STATE Sample $\s_{N,2} \thicksim U(\max(0,\frac{1}{2}(b_u-1)), \frac{1}{3}b_u)$
    \STATE Set $\s_{N,1}\gets b_u-2\s_{N,2}$
    \STATE Update $\Sigma_{N,1}\gets \s_{N,1}$ and $\Sigma_{N,2}\gets \s_{N,2}$
    \RETURN $\Sigma$
\end{algorithmic}
\end{algorithm}

\begin{algorithm}[t]
\small
\caption{\texttt{sample-system($N$,$b$)}: Sample a system composed of $N$ 2D affine transforms $\{(A_k, \mathbf{b}_k) : k=1,\dots, N\}$.}
\begin{algorithmic}[1] \label{alg:sample-system}
    \renewcommand{\algorithmicrequire}{\textbf{Input:}}
    \renewcommand{\algorithmicensure}{\textbf{Output:}}
    \REQUIRE $N \ge 2$, the size of the system; and $b$, a bound on the values of $\mathbf{b}_k$ such that $-b \le \mathbf{b}_{k,i} \le b$
    \ENSURE A set of $N$ affine transformation parameters $(A_k, \mathbf{b}_k)$
    \STATE \textbf{Initialize:} $S\gets \{\}$, empty set of transforms
    \STATE Sample $\alpha \thicksim U(\frac{1}{2}(5+N), \frac{1}{2}(6+N))$
    \STATE $\Sigma \gets$\texttt{sample-svs($N$,$\alpha$)}, $N\times2$ array of singular values
    \FOR{$k=1$ to $N$}
        \STATE Sample $\theta_k,\phi_k \thicksim U(-\pi, \pi)$
        \STATE Sample $d_{k,1},d_{k,2} \thicksim U(\{-1,1\})$
        \STATE Sample $b_{k,1},b_{k,2} \thicksim U(-b, b)$
        \STATE $R_{\theta_k}\gets \bmat{\cos\theta_k & -\sin\theta_k \\ \sin\theta_k & \cos\theta_k}$
        % \STATE $R_{\phi_k}\gets \bmat{d_{k,1}\cos\phi_k & -d_{k,2}\sin\phi_k \\ d_{k,1}\sin\phi_k & d_{k,2}\cos\phi_k }$
        \STATE $R_{\phi_k}\gets \bmat{\cos\phi_k & -\sin\phi_k \\ \sin\phi_k & \cos\phi_k }$
        \STATE $A_k \gets R_{\theta_k} \bmat{\Sigma_{k,1} & 0 \\ 0 & \Sigma_{k,2}} R_{\phi_k} \bmat{d_{k,1} & 0 \\ 0 & d_{k,2}}$
        \STATE $\mathbf{b}_k \gets \bmat{b_{k,1}\\ b_{k,2}}$
        \STATE Insert $(A_k, \mathbf{b}_k)$ into $S$
    \ENDFOR
    \RETURN $S$
\end{algorithmic}
\end{algorithm}

\section{Fractal Pre-training Images}

Here we provide additional details on the proposed fractal pre-training images, including details on how the images are rendered as well as our procedures for ``just-in-time`` (on-the-fly) image generation during training.

\subsection{Rendering Details}\label{sec:rendering-details}

In order to add additional diversity to the rendered fractal images---to encourage the neural network to learn better, more robust representations---we supplement the rendering process (described in Section~\ref{sec:render-fractals}) in three ways. First, we follow the example of~\cite{Kataoka2020Pre-trainingImages} and apply patch-based rendering, which was shown to perform better than simple point rendering. Second, we color the points on the fractal instead of rendering them as grayscale. And third, we add randomly generated backgrounds. Fig.~\ref{fig:rendered-fractals} shows an example rendered fractal image with these properties (far right).

% \noindent\textbf{Patch-based Rendering}\quad
\paragraph{Patch-based Rendering}
Instead of mapping each point in $\iterpoints$ to a single pixel, we follow the approach taken in~\cite{Kataoka2020Pre-trainingImages} and map each point to a patch centered on that pixel. For each image, a patch is sampled uniformly from the set of $3\times 3$ binary patches $\{0,1\}^{3\times 3}$. This patch is applied for each point in $\iterpoints$.

\begin{figure}[t]
    \centering
    \includegraphics[width=\linewidth]{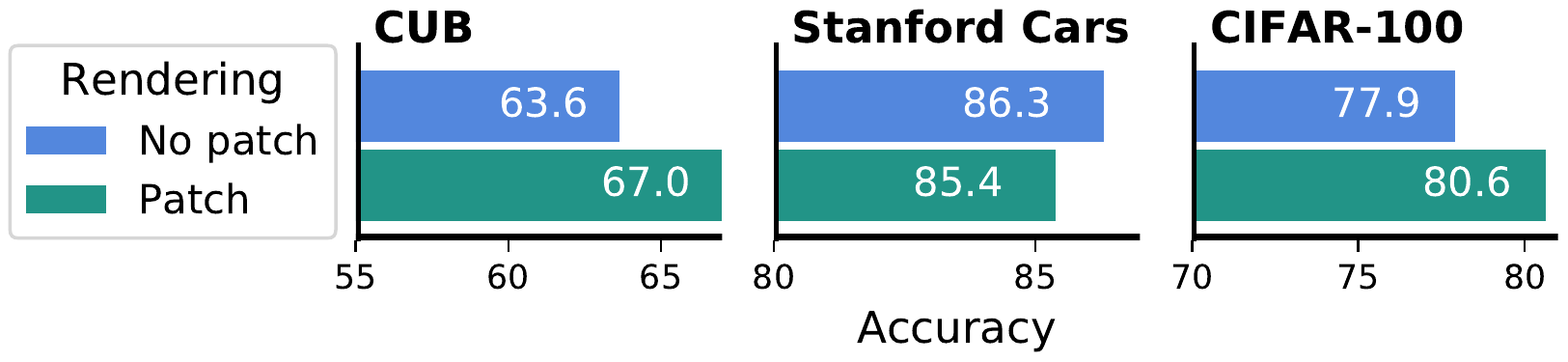}
    \caption{\small Fine-tuning results using models pre-trained with or without patch-based rendering.}
    \label{fig:patch-rendering-results}
\end{figure}

%%%%%%%%%%%%%%%%%%%%%%%%%%%%%%%%%%%%%%%
% First layer Filters Figure
\begin{figure*}[ht]
    \centering
    \includegraphics[width=\linewidth]{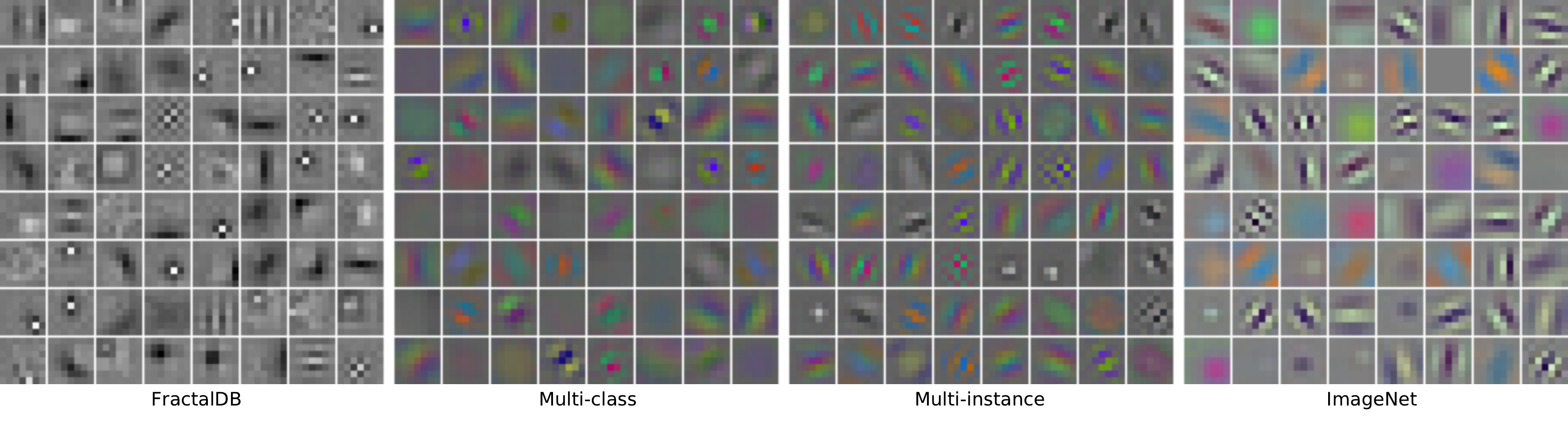}
    \caption{\small First layer filters learned by different pre-training methods.}
    \label{fig:first-layer-filters}
\end{figure*}
%%%%%%%%%%%%%%%%%%%%%%%%%%%%%%%%%%%%%%%

\noindent\textbf{Note:} Kataoka~\etal~\cite{Kataoka2020Pre-trainingImages} found that patch rendering provided a fairly significant performance boost to fine-tuning. We trained a model without patch-based rendering in order to validate their findings---the results are shown in Figure~\ref{fig:patch-rendering-results}. Our findings are consistent with~\cite{Kataoka2020Pre-trainingImages}, although for Stanford Cars the results were slightly better without patch-rendering for some reason.

% \noindent\textbf{Colored Fractals}\quad
\paragraph{Colored Fractals}
We adopt a simple approach for randomly coloring a fractal. First, we render the fractal in grayscale, using density-based rendering (instead of binary). Then we choose a random reference hue value, $h$, and assign a hue to each pixel by treating its (normalized density) grayscale value as an offset from $h$. We randomly sample saturation $s\thicksim U(0.3, 1)$ and value $v\thicksim U(0.5, 1)$
% \footnote{Throughout the paper, we use $U(a, b)$ to mean a continuous uniform distribution on the interval $[a, b]$, and $U(\{\cdot\})$ to mean a discrete uniform distribution over elements of the set $\{\cdot\}$.}
and apply them globally to each pixel to get an HSV image $X^{hsv}$, where the color for pixel $i$ is set to be $X^{hsv}_i\! = ((h+X_i)\pmod{256}, s, v)$. We then convert $X^{hsv}$ to its RGB representation $X^{rgb}$.

% \noindent\textbf{Random Backgrounds}\quad
\paragraph{Random Backgrounds}
Adding backgrounds to the fractal images increases the diversity of images, and should cause the neural network model to learn to ignore backgrounds when making classification decisions. We use the midpoint-displacement, or ``diamond-square'' algorithm~\cite{Fournier1982ComputerModels}, to efficiently generate background textures. A parameter $\gamma$ controls the roughness of the resulting texture. To generate a background, we first sample $\gamma \thicksim U(0.4, 0.8)$ and generate a grayscale texture image using the diamond-square algorithm. Then we colorize the texture using a process similar to the one previously described for colorizing the fractals. The final image is formed by compositing the colored fractal image on top of the random background.

\subsection{Just-In-Time Image Generation} \label{sec:jit-image-generation}

With the correct procedure, we are able to \emph{generate} all images ``on the fly'' as they're needed for training. This is significant, as we circumvent the typical need to store or transmit a huge quantity of data. The entire dataset can be generated from the set of IFS codes, which can be stored in tens or hundreds of megabytes (depending on the number and size of the systems). 
%For reference, a file consisting of one million IFS codes evenly distributed in size between 2 and 4 requires less than 200MB of disk space.  
For context, the ILSVRC2012 subset of ImageNet that is typically used for pre-training comprises 1.281M images and occupies 150GB of disk space.  While in practice, we use dozens of systems per class and their augmentations (approximately 7.2MB for 1000 classes), even if 1.28M images were stored systematically as unique IFS parameters on disk, that only occupies 184.5MB, an $800\times$ reduction in storage.

Three things are necessary in order for image generation to keep up with model throughput: the first is compute-efficient fractal images; the second is efficient code; and the third is retaining a cache of recently-computed objects. 
Affine Iterated Function Systems are computationally efficient---a good approximation of the attractor can be achieved with a few tens or hundreds of thousands of iterations, and don't require any operations beyond basic arithmetic. We are able to get highly-efficient code by carefully writing our algorithms and compiling them with Numba~\cite{Lam2015Numba:Compiler}.

Even with fast code and efficient fractals, it may not be possible to generate images fast enough to match model throughput, particularly when training on multiple GPUs. As a solution, we keep a cache of recently-computed fractal images, which gets updated on a fixed schedule. For example, when training a multi-class classification model, we keep a cache of the last 512 generated images. Half of each training batch consists of images drawn from the cache and augmented using standard data augmentation practices. The other half of the batch consists of newly-generated images, which are then used to update the cache. This cuts in half the number of images that need to be generated from scratch at each iteration of training, greatly easing the computational load. Using a cache is even more critical when generating multi-instance images, as we describe in Section~\ref{sec:multi-instance-images}.

\noindent \textbf{Note:} Our target in this work is to generate images fast enough to keep up with training a ResNet50 model using distributed training on a workstation with 8 GPUs. Different hardware setups and different models may require adjustments---such as different cache sizes or update intervals---but with proper tuning the approach should work in a wide variety of circumstances.

\begin{figure}
    \centering
    \includegraphics[width=\linewidth]{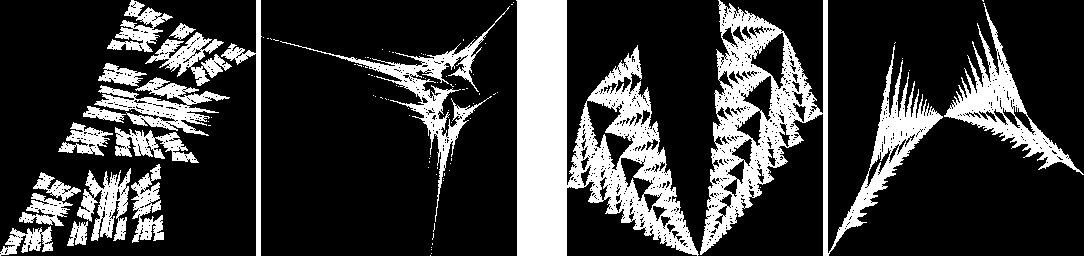}
    \caption{\small Examples of the effect that small perturbations in parameters can have on the resulting fractal images. In each of the two examples shown, the value of a single parameter in the IFS code was shifted by $0.1$.}
    \label{fig:small-param-diff}
\end{figure}

\begin{table*}[t]
    \label{tab:sampling-times}
    \centering
    {
    \renewcommand{\arraystretch}{1.2}
    \begin{tabular}{l|ccccccc}
        \hline \hline
        \multicolumn{1}{c|}{} & \multicolumn{7}{c}{$N$} \\
        \cline{2-8}
        & 2 & 3 & 4 & 5 & 6 & 7 & 8 \\ 
        \hline
        Sample-svs & $11.7 \pm 0.49$ & $17.3 \pm 0.76$ & $22.4 \pm 0.28$ & $28.2 \pm 0.38$ & $33.1 \pm 0.17$ & $38.3 \pm 0.48$ & $43.3 \pm 0.43$ \\
        Sample-system & $42.8 \pm 1.04$ & $49.2 \pm 0.75$ & $55.4 \pm 0.25$ & $60.5 \pm 0.40$ & $67.0 \pm 0.43$ & $72.7 \pm 0.55$ & $80.7 \pm 2.86$ \\ 
        \hline \hline
    \end{tabular}}
    \vspace{1mm}
    \caption{Average time (in microseconds) for sampling IFS codes of different size ($N$), using our Python implementation. The first row shows times for sampling singular values alone, and the second row shows times for sampling the full system (including sampling singular values).}
\end{table*}

\section{Computational Requirements}

\noindent \textbf{Fractal Sampling and Rendering}\quad For reference, we report compute time for sampling systems and rendering fractal images. Compute time was measured using an Intel Xeon E3-1245 3.7GHz CPU. In Table~\ref{tab:sampling-times}, we report the average time for sampling IFS codes for systems of size $2$ up to size $8$, along with the time for sampling just the singular values. In Table~\ref{tab:rendering-times}, we report the average time required for various stages of the image rendering process. The memory requirements for rendering a single image are low, requiring little more than the size of the output array.

\begin{table}[]
    \centering
    {
    \renewcommand{\arraystretch}{1.2}
    \begin{tabular}{
        m{0.5\linewidth}|>{\centering\arraybackslash}m{0.375\linewidth}
    }
        \hline \hline
        Operation & Time (ms) \\ 
        \hline
        Iterate ($10^5$) & $4.39 \pm 0.034$ \\ 
        Render ($256\times 256$) & $1.46 \pm 0.017$ \\ 
        Colorize & $0.23 \pm 0.002$ \\
        Background ($256 \times 256$) & $0.77 \pm 0.001$ \\
        \hline \hline
    \end{tabular}}
    \vspace{1mm}
    \caption{Average time (in milliseconds) for various stages of the fractal image rendering process, using our implementation (Python and Numba~\cite{Lam2015Numba:Compiler}). (\textbf{Iterate}) produces coordinates on the attractor through random iteration (100,000 iterations); (\textbf{Render}) maps the coordinates to a $256\times 256$ grayscale image using patch-based rendering; (\textbf{Colorize}) converts the grayscale image to a color image; (\textbf{Background}) renders a random background. See \ref{sec:rendering-details} for details.}
    \label{tab:rendering-times}
\end{table}

\noindent \textbf{Training}\quad In Table~\ref{tab:training-times}, we report training time under two different hardware settings. The first is a single node with 8 1080-Ti GPUs and 48 CPU cores. The second is two nodes, each with 4 Tesla P100 GPUs and a total of 56 CPU cores. We report the time required to train a model for 90 epochs, or 90,000,000 iterations (1,000,000 images per epoch, comparable to ILSVRC 2012), for both single-instance multi-class classification and multi-instance prediction.

\begin{table}[]
    \centering
    {
    \renewcommand{\arraystretch}{1.2}
    \begin{tabular}{
        m{0.3\linewidth}|>{\centering\arraybackslash}m{0.265\linewidth}>{\centering\arraybackslash}m{0.265\linewidth}
    }
        \hline \hline
        Task & {\small$1\!\times\!8$ 1080-Ti} & {\small$2\!\times\!4$ P100} \\ 
        \hline
        Multi-class & $23$h ($15.3$m) & $18$h ($12$m) \\
        Multi-instance & $25$h ($16.6$m) & $19.5$h ($13$m) \\
        \hline \hline
    \end{tabular}}
    \vspace{1mm}
    \caption{Representative pre-training times for both multi-class classification and multi-instance prediction, for two different hardware stacks: one node with 8 1080-Ti GPUs, and two nodes with 4 P100 GPUs each. The time in hours to train for 90 epochs is shown, with the approximate per-epoch training time (in minutes) shown in parentheses.}
    \label{tab:training-times}
\end{table}

\section{Additional Data Examples} \label{sec:data-examples}

\subsection{Small Changes to Parameters}
Section~\ref{sec:multi-class-classification} pointed out that small perturbations to IFS codes can sometimes result in large visual differences in the corresponding fractal images. We show examples of this in Figure~\ref{fig:small-param-diff}.

\subsection{Problems in FractalDB}

\begin{figure}
    \centering
    \includegraphics[width=\linewidth]{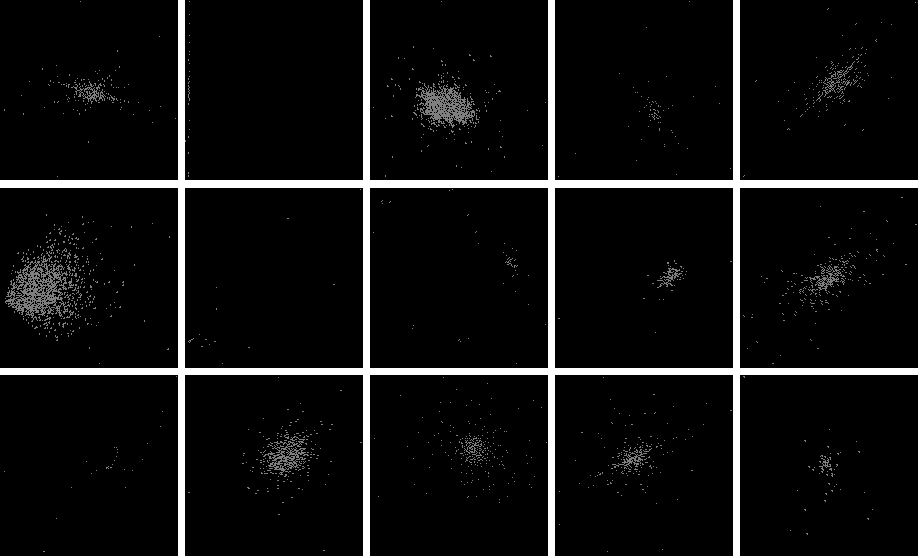}
    \caption{\small Examples of degenerate FractalDB images, caused by IFS parameter augmentation leading to non-contractive systems.}
    \label{fig:fracdb-bad-examples}
\end{figure}

Since the data augmentation process used for FractalDB~\cite{Kataoka2020Pre-trainingImages} doesn't enforce contractivity in the resulting IFS codes, some of the resulting images are degenerate. Figure~\ref{fig:fracdb-bad-examples} shows some sample images from the FractalDB dataset that exhibit this degeneracy, leading to small clouds of points or mostly empty images.

\subsection{Example Images}

\begin{figure*}
    \centering
    \includegraphics[width=\linewidth]{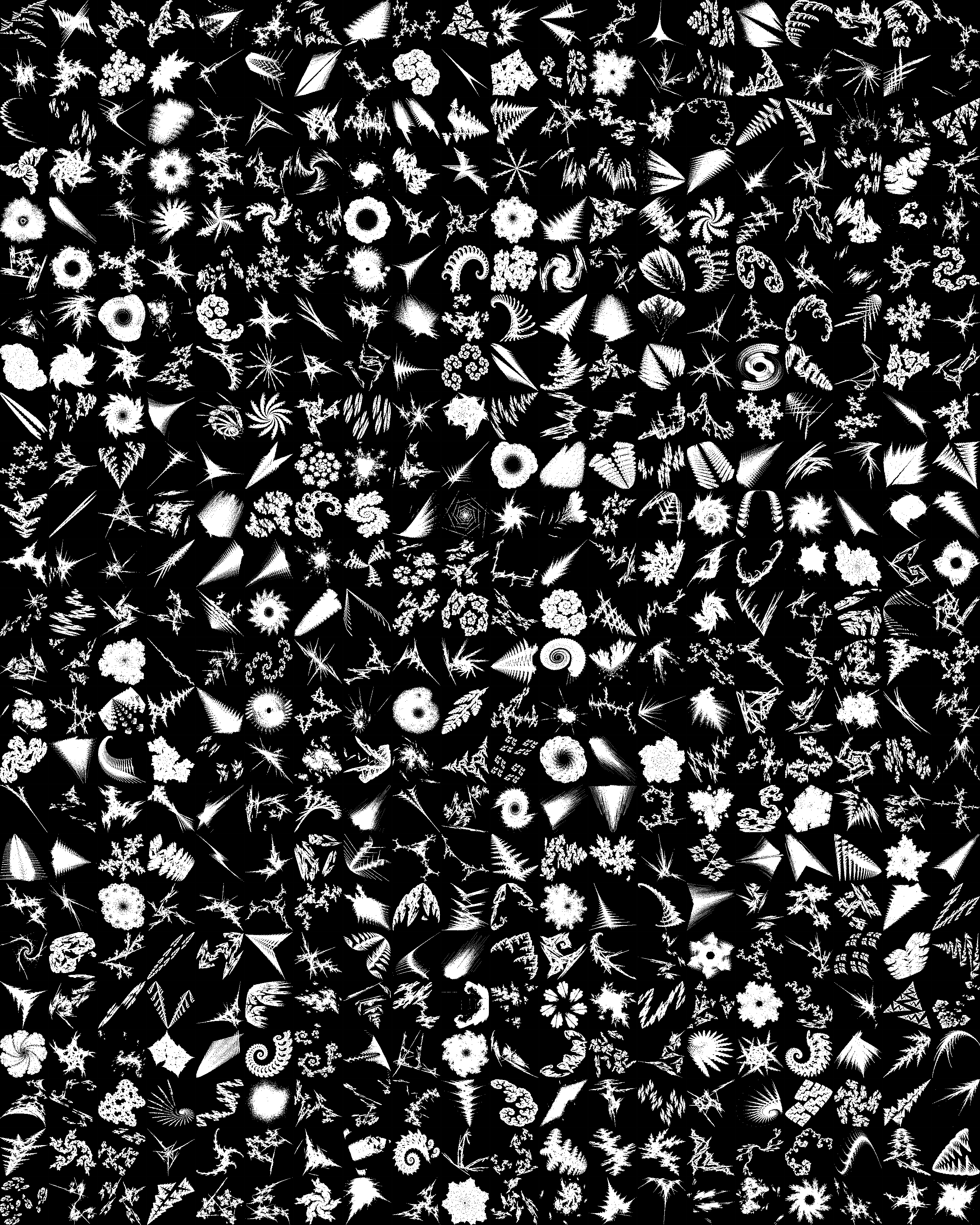}
    \caption{\small Example images from 500 different systems used in our fractal pre-training.}
    \label{fig:example-images}
\end{figure*}

Figure~\ref{fig:example-images} shows images of 500 (out of 50,000) Iterated Function Systems sampled according to Algorithms~\ref{alg:sample-svs} and~\ref{alg:sample-system}, and used to pre-train the models for which we report results in the paper. We show just the binary-rendered fractal images (without color or background) to give a clear picture of the fractal geometry.

\subsection{System Probabilities $p_i$} \label{sec:system-probabilities}

\begin{figure}[t]
    \centering
    \includegraphics[width=\linewidth]{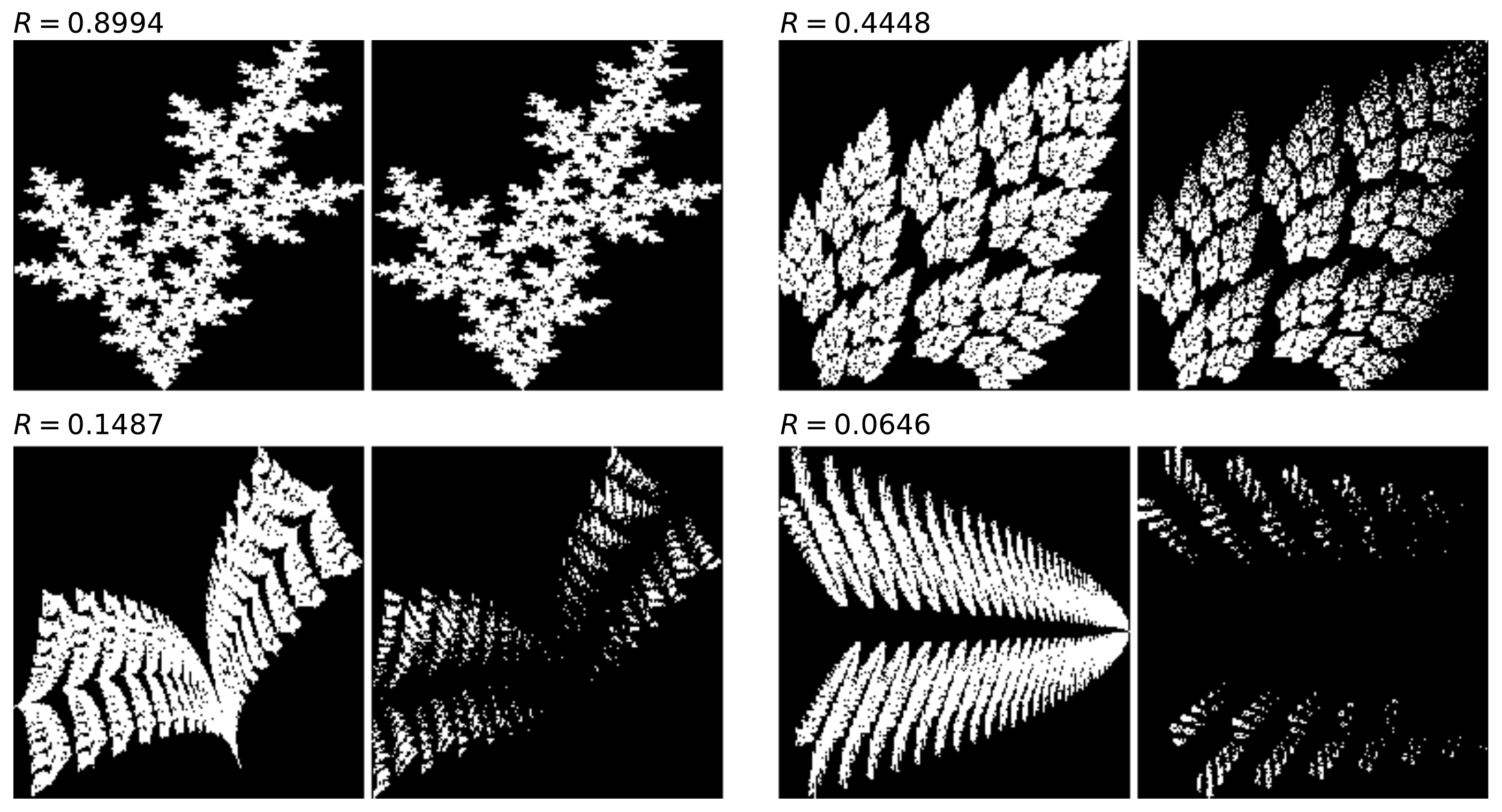}
    \caption{Rendered IFS codes using different probabilities (determinant-based on the left, uniform on the right of each pair). Uniform probabilities don't work well when the determinants of the system have significantly different magnitudes. $R$ is the ratio of the smaller to the larger determinant.}
    \label{fig:probability-comparison}
\end{figure}

An affine IFS code consists of a set of affine functions, each with an associated sampling probability (see Section~\ref{sec:fractal-images}). The sampling probabilities $p_i$ don't affect the shape of the underlying attractor, but they do influence the distribution of points on the attractor that are visited during iteration. Figure~\ref{fig:probability-comparison} shows several IFS rendered using two different choices for $p_i$: (1) $p_i$ is proportional to the magnitude of the determinant of the linear part of the transform, $p_i \propto \left|\det A_i\right|$; and (2) $p_i$ is uniform, $p_i = \frac{1}{\left|S\right|}$. When one determinant is significantly larger than the other, there are parts of the attractor that don't get visited during iteration using uniform $p_i$. We use the determinant method for setting $p_i$ in all our experiments.

\section{First Layer Filters}

In Figure~\ref{fig:first-layer-filters}, we show a comparison of the filters from the first layer of ResNet50, pre-trained using different methods. Interestingly, it appears that filters learned from multi-instance prediction are closest to those learned by pre-training on ImageNet.

\section{Additional Results}

\begin{figure}[t]
    \centering
    \includegraphics[width=\linewidth]{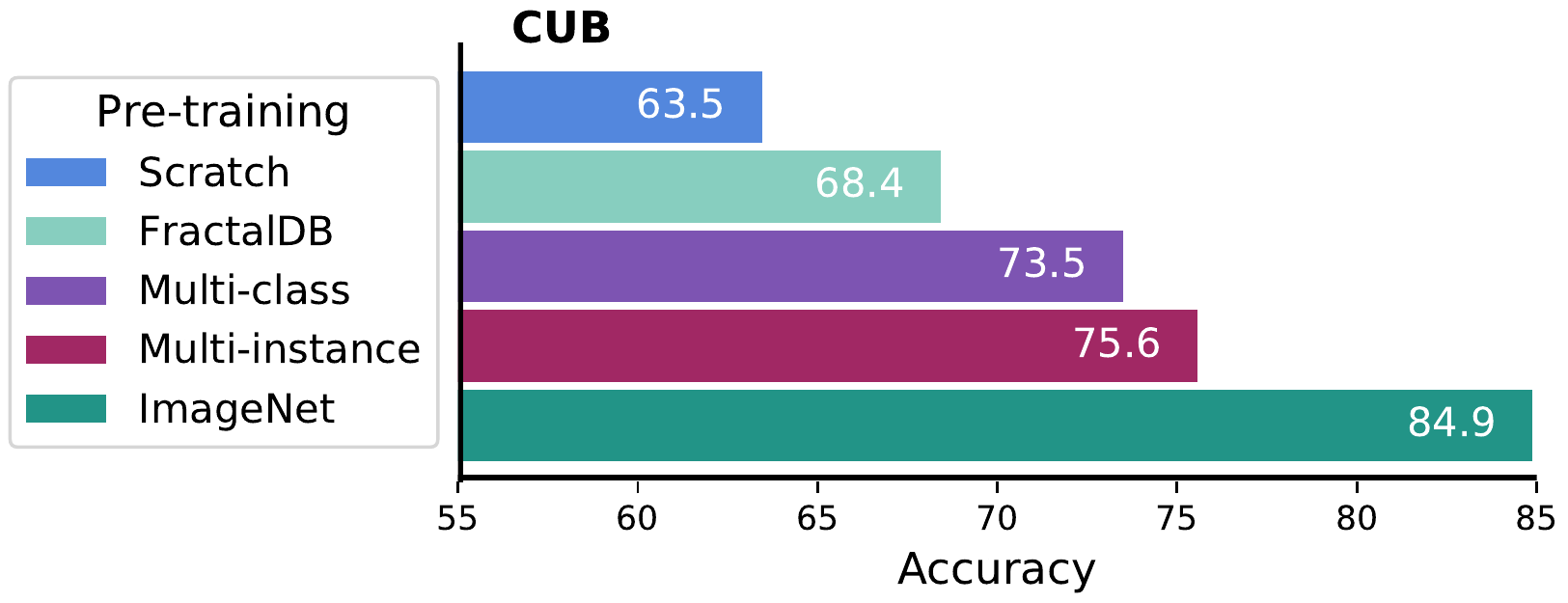}
    \caption{\small Results of fine-tuning on CUB using a larger image resolution ($448\times 448$). The pre-trained networks are the same as in Figure~\ref{fig:finetune-results}.}
    \label{fig:cub-448-results}
\end{figure}

Here we include some additional experimental results that didn't fit in the main body of the paper. Our main set of experiments evaluated fine-tuning performance using image resolution $224\times 224$. One common way to achieve better performance is to use a larger image resolution, such as $448 \times 448$. We fine-tuned on CUB using this resolution, and the results are shown in Figure~\ref{fig:cub-448-results}. We see better performance across the board, with FractalDB now outperforming training-from-scratch, and with the relative performance order otherwise staying the same. At the higher resolution, we also see the gap between ImageNet and fractal pre-training get wider, indicating there is still plenty of work to do to improve the fractal pre-training methods.

\begin{figure}[t]
    \centering
    \includegraphics[width=\linewidth]{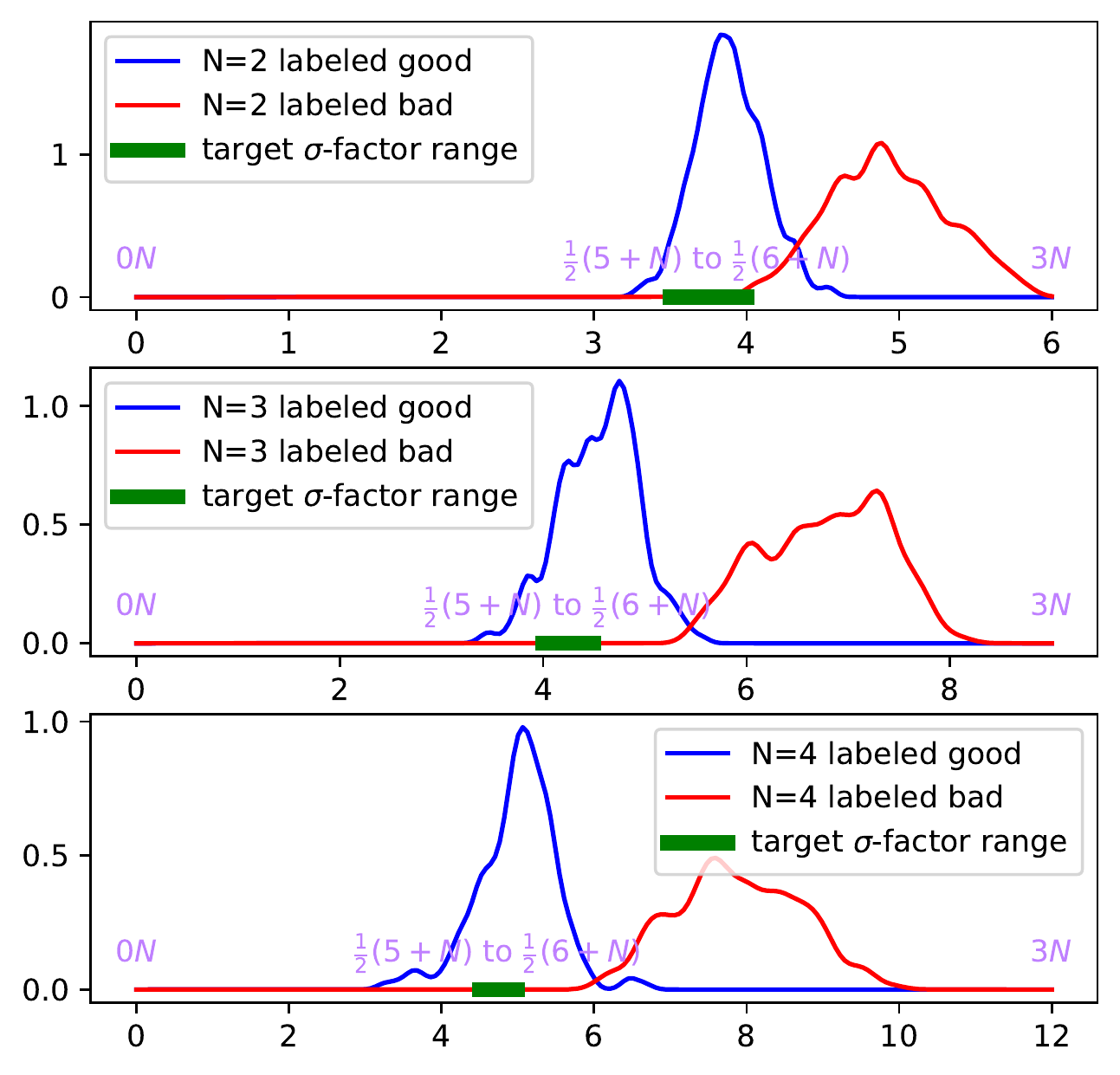}
    \caption{\small Plot of \sfactor\ densities of hand-labeled systems with $N=2,3,4$. It is important to note that the $\frac{1}{2} \left(5+N\right)$ to $\frac{1}{2} \left(6+N\right)$ range was chosen well before these plots were generated; the plots strongly validate the selected range.}
    \label{fig:labeled-densities}
\end{figure}

\section{\sfactor\ Density for Hand-labeled Systems}

In Figure \ref{fig:labeled-densities}, we show the distribution of \sfactor s for the hand-labeled systems discussed in Section \ref{sec:sampling-algos}.  For each value of $N \in \{2,3,4\}$, several hundred systems were labeled as to whether or not they had subjectively ``good'' geometry.  It is critical to point out here that these plots were generated only \emph{after} the range of 
 $\frac{1}{2} \left(5+N\right)$ to $\frac{1}{2} \left(6+N\right)$ was determined empirically; however, the plots  strongly validate the range selected.

\null
\vfill

\end{document}